\documentclass[10pt,twocolumn,letterpaper]{article}

\usepackage[pagenumbers]{cvpr} 

\usepackage[dvipsnames]{xcolor}

\definecolor{cvprblue}{rgb}{0.21,0.49,0.74}
\usepackage[pagebackref,breaklinks,colorlinks,citecolor=cvprblue]{hyperref}
\usepackage{float}
\usepackage{multirow}
\usepackage{makecell}
\usepackage{algorithm}
\usepackage{algpseudocode}
\usepackage{pifont}
\usepackage{tablefootnote}
\usepackage[accsupp]{axessibility}

\def\eg{\emph{e.g.},\ } 
\def\ie{\emph{i.e.},\ }

\newcommand{\bi}[1]{\textcolor{black}{#1}}
\newcommand{\cam}[1]{\textcolor{black}{#1}}

\def\paperID{4751} 
\def\confName{CVPR}
\def\confYear{2024}

\title{PartDistill: 3D Shape Part Segmentation by Vision-Language Model Distillation}

\author{Ardian Umam$^{1}$ \quad Cheng-Kun Yang$^{2}$ \quad Min-Hung Chen$^{3}$ \quad Jen-Hui Chuang$^{1}$ \quad Yen-Yu Lin$^{1}$ \vspace{0.3em} \\
$^1$National Yang Ming Chiao Tung University \quad $^2$MediaTek \quad $^3$NVIDIA
}

\begin{document}
\maketitle
\begin{abstract}
This paper proposes a cross-modal distillation framework, PartDistill, which transfers 2D knowledge from vision-language models (VLMs) to facilitate 3D shape part segmentation. 
PartDistill addresses three major challenges in this task: the lack of 3D segmentation in invisible or undetected regions in the 2D projections, inconsistent 2D predictions by VLMs, and the lack of knowledge accumulation across different 3D shapes. 
PartDistill consists of a teacher network that uses a VLM to make 2D predictions and a student network that learns from the 2D predictions while extracting geometrical features from multiple 3D shapes to carry out 3D part segmentation. 
A bi-directional distillation, including forward and backward distillations, is carried out within the framework, where the former forward distills the 2D predictions to the student network, and the latter improves the quality of the 2D predictions, which subsequently enhances the final 3D segmentation. 
Moreover, PartDistill can exploit generative models that facilitate effortless 3D shape creation for generating knowledge sources to be distilled. 
Through extensive experiments, PartDistill boosts the existing methods with substantial margins on widely used ShapeNetPart and PartNetE datasets, by more than 15\% and 12\% higher mIoU scores, respectively.  The code for this work is available at \href{https://github.com/ardianumam/PartDistill}{https://github.com/ardianumam/PartDistill}.
\end{abstract}    
\vspace{-10px}
\section{Introduction}
\label{sec:intro}
\vspace{-5px}
3D shape part segmentation is essential to various 3D vision applications, such as shape editing~\cite{neuform_2022,shapeeding_2021}, stylization~\cite{text2mesh_2022}, and augmentation~\cite{pointmixswap_2022}. 
Despite its significance, acquiring part annotations for 3D data, such as point clouds or mesh shapes, is labor-intensive and time-consuming.

Zero-shot learning~\cite{zeroshot_survey2019,zeroshot_pointcloud_cls_2022} generalizes a model to unseen categories without annotations and has been notably uplifted by recent advances in vision-language models (VLMs)~\cite{clip_2021,glipv1_2022,glipv2_2022,lseg_2022}. 
%
By learning on large-scale image-text data pairs, VLMs show promising generalization abilities on various 2D recognition tasks.
Recent research efforts~\cite{pointclipv1_2022, pointclipv2_2023, partslip_2023, satr_2023} have been made to utilize VLMs for zero-shot 3D part segmentation, where a 3D shape is projected into multi-view 2D images, and a VLM is applied to these images for 2D prediction acquisition.
Specifically, PointCLIP~\cite{pointclipv1_2022} and PointCLIPv2~\cite{pointclipv2_2023} produce 3D point-wise semantic segmentation by averaging their corresponding 2D pixel-wise predictions. 
Meanwhile, PartSLIP~\cite{partslip_2023} and SATR~\cite{satr_2023} present a designated weighting mechanism to aggregate multi-view bounding box predictions.

\begin{figure}[t]
  \centering
  \includegraphics[keepaspectratio, width=0.95\columnwidth]{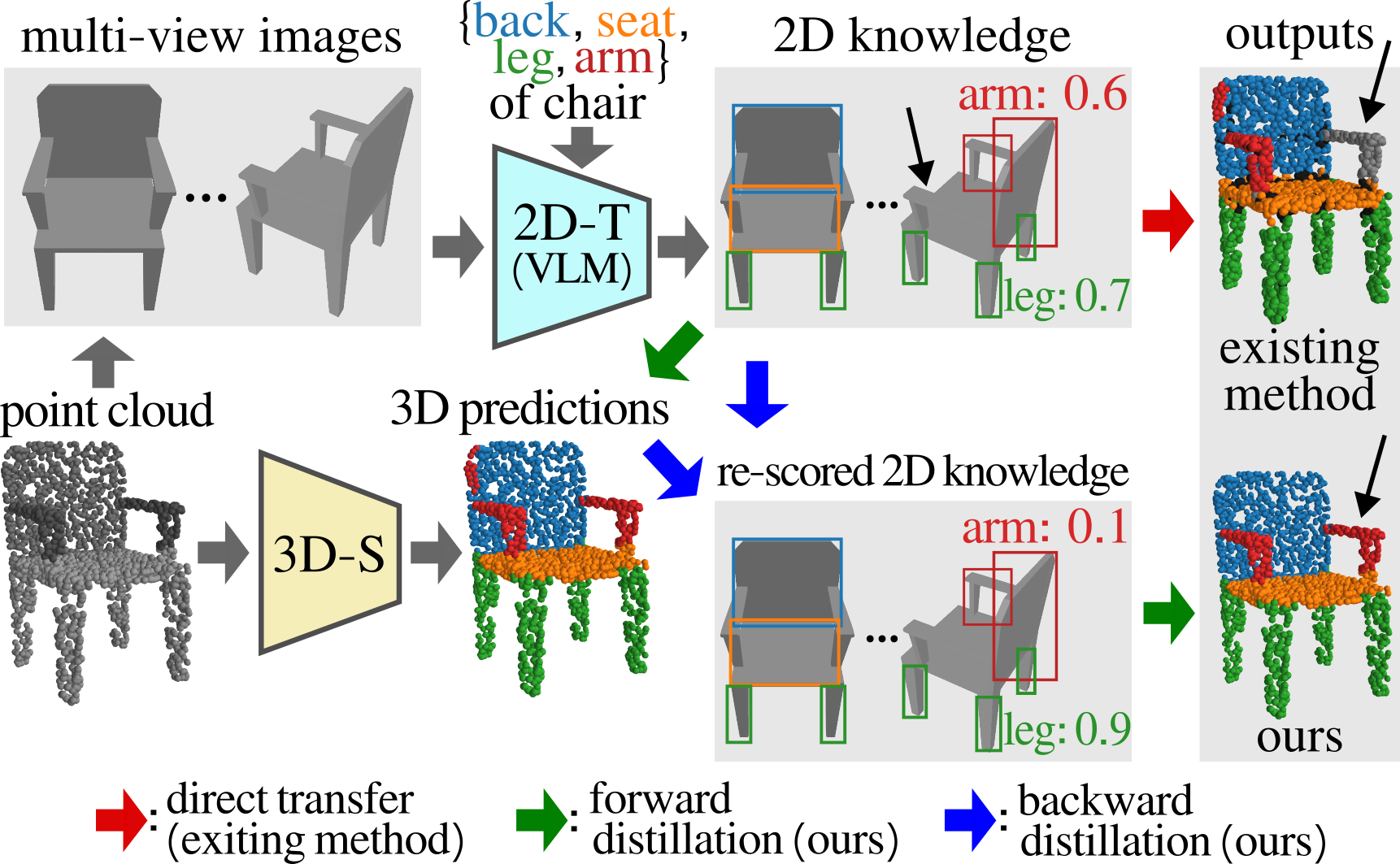}
  \caption{
  We present a distillation method that carries out zero-shot 3D shape part segmentation with a 2D vision-language model.
  After projecting an input 3D point cloud into multi-view 2D images, the 2D teacher (2D-T) and the 3D student (3D-S) networks are applied to the 2D images and 3D point cloud, respectively.
  Instead of direct transfer, our method carries bi-directional distillations, including forward and backward distillations, and yields better 3D part segmentation than the existing method.
  \vspace{-13px}
  }
  \label{fig:teaser}
\end{figure}

The key step of zero-shot 3D part segmentation with 2D VLMs, \eg \cite{pointclipv1_2022,pointclipv2_2023,partslip_2023,satr_2023}, lies in the transfer from 2D pixel-wise or bounding-box-wise predictions to 3D point segmentation. 
This step is challenging due to three major issues. 
First ($\boldsymbol{\mathcal{I}_1}$), 
some 3D regions lack corresponding 2D predictions in multi-view images, which are caused by occlusion or not being covered by any bounding boxes, illustrated with black and gray points, respectively, in Fig.~\ref{fig:teaser}. 
This issue is considered a limitation in the previous work~\cite{pointclipv1_2022,pointclipv2_2023,partslip_2023,satr_2023}. 
Second ($\boldsymbol{\mathcal{I}_2}$), there exists potential inconsistency among 2D predictions in multi-view images caused by inaccurate VLM predictions.
%
%
Third ($\boldsymbol{\mathcal{I}_3}$), existing work~\cite{pointclipv1_2022,pointclipv2_2023,partslip_2023,satr_2023} directly transfers 2D predictions to segmentation of a single 3D shape. 
The 2D predictions yielded based on appearance features are not optimal for 3D geometric shape segmentation, while geometric evidence given across different 3D shapes is not explored.

To alleviate the three issues $\boldsymbol{\mathcal{I}_1}$$\sim$$\boldsymbol{\mathcal{I}_3}$, unlike existing methods~\cite{pointclipv1_2022,pointclipv2_2023,partslip_2023,satr_2023} directly transferring 2D predictions to 3D segmentation, we propose a cross-modal distillation framework with a teacher-student model.
%
Specifically, a VLM is utilized as a 2D teacher network, accepting multi-view images of a single 3D shape. 
The VLM is pre-trained on large-scale image-text pairs and can exploit appearance features to make 2D predictions. 
The student network is developed based on a point cloud backbone.
It is derived from multiple, unlabeled 3D shapes and can extract point-specific geometric features.
The proposed distillation method, \emph{\textbf{PartDistill}}, leverages the strengths of both networks, hence improving zero-shot 3D part segmentation.

The student network learns from not only the 2D teacher network but also 3D shapes.
It can extract point-wise features and segment 3D regions uncovered by 2D predictions, hence tackling issue $\boldsymbol{\mathcal{I}_1}$.
As a distillation-based method, PartDistill tolerates inconsistent predictions between the teacher and student networks, which alleviates issue $\boldsymbol{\mathcal{I}_2}$ of negative transfer caused by wrong VLM predictions.
The student network considers both appearance and geometric features.
Thus, it can better predict 3D geometric data and mitigate issue $\boldsymbol{\mathcal{I}_3}$.
As shown in Fig.~\ref{fig:teaser}, the student network can correctly predict the undetected arm of the chair (see the black arrows) by learning from other chairs.

PartDistill carries out a bi-directional distillation.
%
It first \emph{forward distills} the 2D knowledge to the student network.
We observe that after the student integrates the 2D knowledge, we can jointly refer both teacher and student knowledge to perform \emph{backward distillation} which re-scores the 2D knowledge based on its quality. Those of low quality will be suppressed with lower scores, such as from 0.6 to 0.1 for the falsely detected arm box in Fig.~\ref{fig:teaser}, and vice versa.  Finally, this
re-scored knowledge is utilized by the student network to seek better 3D segmentation.

The main contributions of this work are summarized as follows. 
First, we introduce PartDistill, a cross-modal distillation framework that transfers 2D knowledge from VLMs to facilitate 3D part segmentation. 
PartDistill addresses three identified issues present in existing methods and generalizes to both VLM with bounding-box predictions (B-VLM) and pixel-wise predictions (P-VLM).
Second, we propose a bi-directional distillation, which involves enhancing the quality of 2D knowledge and subsequently improving the 3D predictions.
Third, PartDistill can leverage existing generative models~\cite{dit3d_2023,shapeaspoint_2021} to enrich knowledge sources for distillation.
Extensive experiments demonstrate that PartDistill surpasses existing methods by substantial margins on widely used benchmark datasets, ShapeNetPart~\cite{shapenet_2016} and PartNetE~\cite{partslip_2023}, with more than 15\% and 12\% higher mIoU scores, respectively. 
PartDistill consistently outperforms competing methods in zero-shot and few-shot scenarios on 3D data in point clouds or mesh shapes.


\begin{figure*}[t]
  \centering
  \includegraphics[keepaspectratio, width=0.97\textwidth]{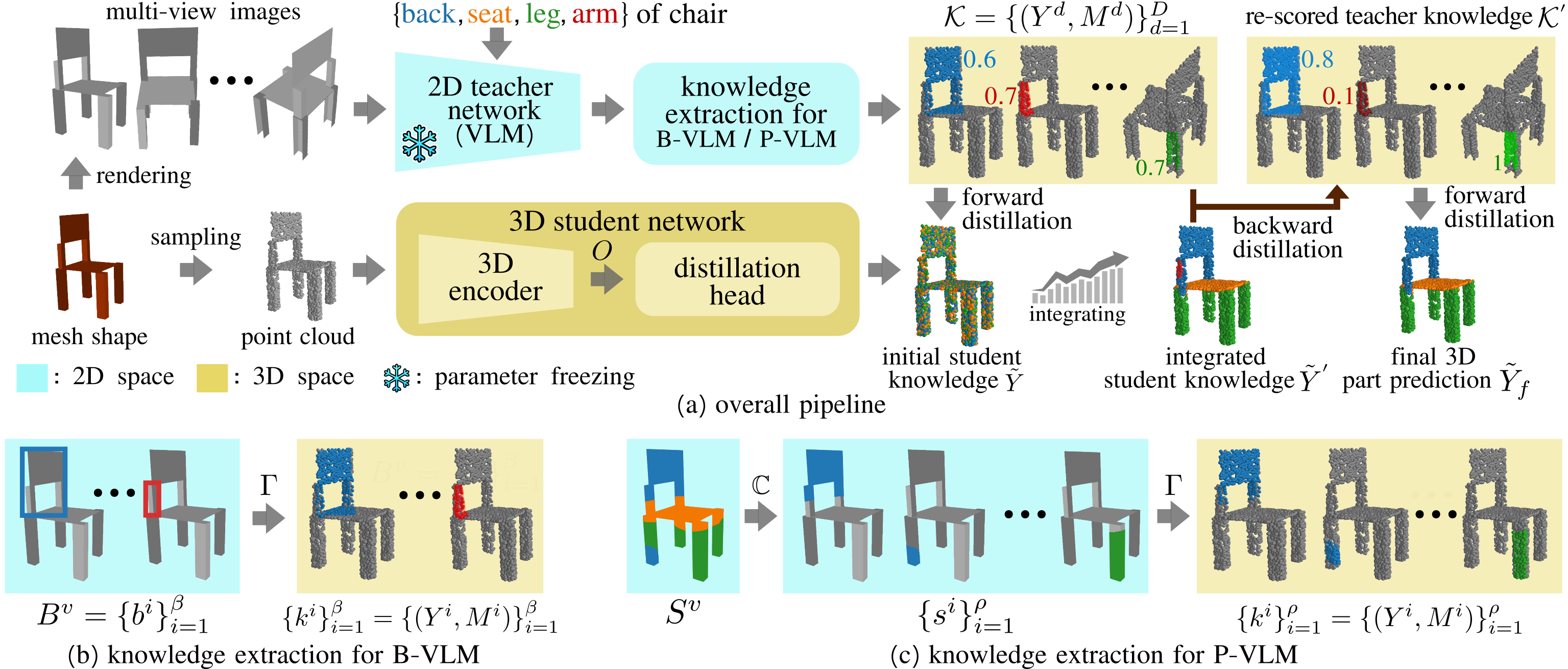}
  \vspace{-5px}
  \caption{\textbf{Overview of the proposed method.} (a) The overall pipeline where the knowledge extracted from a vision-language model (VLM) is distilled to carry out 3D shape part segmentation by teaching a 3D student network. 
  Within the pipeline, \emph{backward distillation} is introduced to re-score the teacher's knowledge based on its quality and subsequently improve the final 3D part prediction. 
  (b) $\&$ (c) Knowledge is extracted by back-projection when we adopt (b) a bounding-box VLM (B-VLM) or (c) a pixel-wise VLM (P-VLM), where $\Gamma$ and $\mathbb{C}$ denote 2D-to-3D back-projection and connected component labeling~\cite{connectedcomp_2019}, respectively.
  \vspace{-10px}}
  \label{fig:main_fig}
\end{figure*}
\vspace{-5px}
\section{Related Work}
\label{sec:related_work}

\vspace{-5px}
\paragraph{Vision-language models.}
%
%
Based on learning granularity, vision-language models (VLMs) can be grouped into three categories, including the image-level~\cite{clip_2021,align_2021}, pixel-level~\cite{lseg_2022,clipseg_2022,seem_2023}, and object-level~\cite{glipv1_2022,glipv2_2022,groundingdino_2023} categories. 
%
%
The second and the third categories make pixel-level and bounding box predictions, respectively, while the first category produces image-level predictions. 
Recent research efforts on VLMs have been made for cross-level predictions.
For example, pixel-level predictions can be derived from an image-level VLM via up-sampling the 2D features into the image dimensions, as shown in PointCLIPv2~\cite{pointclipv2_2023}. 
%
In this work, we propose a cross-modal distillation framework that learns and transfers knowledge from a VLM in the 2D domain to 3D shape part segmentation. 

\vspace{-13px}
\paragraph{3D part segmentation using vision-language models.} State-of-the-art zero-shot 3D part segmentation~\cite{partslip_2023,pointclipv2_2023,satr_2023} is developed by utilizing a VLM and transferring its knowledge in the 2D domain to the 3D space.  
The pioneering work PointCLIP~\cite{pointclipv1_2022} utilizes CLIP~\cite{clip_2021}.
%
%
PointCLIPv2~\cite{pointclipv2_2023} extends PointCLIP by making the projected multi-view images more realistic and proposing LLM-assisted text prompts~\cite{gpt3_2020}, hence producing more reliable CLIP outputs for 3D part segmentation. 

Both PointCLIP and PointCLIPv2 rely on individual pixel predictions in 2D views to get the predictions of the corresponding 3D points, but individual pixel predictions are less unreliable.
PartSLIP~\cite{partslip_2023} suggests to extract superpoints~\cite{superpoints_2018} from the input point cloud. 
Therefore, 3D segmentation is estimated for each superpoint by referring to a set of relevant pixels in 2D views.
PartSLIP uses GLIP~\cite{glipv1_2022} to output bounding boxes and further proposes a weighting mechanism to aggregate multi-view bounding box predictions to yield 3D superpoint predictions.
%
SATR~\cite{satr_2023} shares a similar idea with PartSLIP but handles 3D mesh shapes instead of point clouds. 
%
%

Existing methods~\cite{partslip_2023,pointclipv2_2023,pointclipv1_2022,satr_2023} directly transfer VLM predictions from 2D images into 3D spaces and pose three issues: ($\boldsymbol{\mathcal{I}_1}$) uncovered 3D points,  ($\boldsymbol{\mathcal{I}_2}$) negative transfer, and ($\boldsymbol{\mathcal{I}_3}$) cross-modality predictions, as discussed before. 
We present a distillation-based method to address all three issues and make substantial performance improvements.

\vspace{-13px}
\paragraph{2D to 3D distillation.} 
Seminal work of knowledge distillation~\cite{distillation_2006,distillation_hinton_2015} aims at transferring knowledge from a large model to a small one.
\cam{Subsequent research efforts~\cite{image2lidar_2022,image2point_2023,depthcontrast_2021,seganypoint_2023,yang20232d} adopt this idea of transferring knowledge from a 2D model for 3D understanding. However, these methods require further fine-tuning with labeled data. OpenScene~\cite{openscene_peng2023} and CLIP2Scene~\cite{clip2scene_chen2023} require no fine-tuning and share a similar concept with our method of distilling VLMs for 3D understanding, with ours designed for part segmentation and theirs for indoor/outdoor scene segmentation. The major difference is that our method can enhance the knowledge sources in the 2D modality via the proposed backward distillation. Moreover, our method is generalizable to both P-VLM (pixel-wise VLM) and B-VLM (bounding-box VLM), while their methods are only applicable to P-VLM.}
\section{Proposed Method}
\label{sec:proposed_method}
\subsection{Overview}
\vspace{-5px}

Given a set of 3D shapes, this work aims to segment each \bi{one}
into $R$ semantic parts without \bi{training with}
any part annotations. 
To this end, we propose a cross-modal \bi{bi-directional }distillation framework\bi{, \textit{\textbf{PartDistill}}, which}
transfers 2D knowledge from a VLM to facilitate 3D shape part segmentation. 
As illustrated in Fig.~\ref{fig:main_fig}, our framework takes triplet data as input, including the point cloud of the shape with $N$ 3D points, multi-view rendered images from the shape in $V$ different poses, and $R$ text prompts with each describing the target semantic part\bi{s} within the 3D shapes.

For the 2D modality, the $V$ multi-view images and the text prompts are fed into a \bi{Bounding-box VLM (}B-VLM\bi{)} or \bi{Pixel-wise VLM (}P-VLM\bi{)}.
For each view $v$, a B-VLM produces a set of bounding boxes, 
$B^v=\{b^i\}_{i=1}^{\beta}$
\bi{while} a P-VLM generates pixel-wise predictions $S^v$. 
We \bi{then} perform \bi{\textit{knowledge extraction}~(Sec~\ref{sec:distill2dto3d})}
for each $B^v$ or $S^v$; 
Namely, we transfer the 2D predictions into the 3D space through back-projection for a B-VLM or connected-component labeling~\cite{connectedcomp_2019} followed by back-projection for a P-VLM, as shown in Fig.~\ref{fig:main_fig}\bi{~(}b\bi{)} and Fig.~\ref{fig:main_fig}\bi{~(}c\bi{)}, respectively.
Subsequently, a set of $D$ \bi{\textit{teacher}} knowledge units, 
$\mathcal{K}=\{k\}_{d=1}^{D}=\{Y^d,M^d\}_{d=1}^{D}$, 
is obtained by aggregating from all $V$ multi-view images. 
Each unit $d$ comprises point-wise part probabilities, $Y^d \in \mathbb{R}^{N \times R}$, from the teacher \bi{VLM} network, accompanied with a mask, $M^d \in \{0,1\}^{N}$, identifying the points included in this knowledge unit.


For the 3D modality, the point cloud is passed into the 3D \bi{\textit{student}} network with a 3D encoder and a distillation head, producing point-wise part prediction\bi{s}, $\tilde{Y} \in \mathbb{R}^{N \times R}$. 
\bi{With the proposed bi-directional distillation framework, we first \textit{forward distill}}
teacher's \bi{2D} knowledge by aligning $\tilde{Y}$ with $\mathcal{K}$ via minimizing the proposed loss, $\mathcal{L}_{distill}$, specified in Sec~\ref{sec:distill2dto3d}. 
The \bi{3D} student network integrates 2D knowledge from the teacher through optimization. 
The integrated student knowledge $\tilde{Y}'$ and the teacher knowledge $\mathcal{K}$ are \bi{then} jointly referred to perform \bi{\textit{backward distillation} from 3D to 2D,}
detailed in Sec.~\ref{subsec:knowledge_refinement}, which re-scores each knowledge unit $k^d$ based on its qualities, as shown in Fig.~\ref{fig:main_fig}.
%
%
\bi{Finally, the re-scored knowledge $\mathcal{K}'$ is used to refine the student knowledge to get final part segmentation predictions $\tilde{Y}_f$ by assigning each point}
%
to the part with the highest probability.    

\subsection{\bi{Forward distillation: 2D to 3D}}
\label{sec:distill2dto3d}
Our method extracts the teacher's knowledge in the 2D modality and distills it in the 3D space. 
In the 2D modality, $V$ multi-view images  $\{I^v \in \mathbb{R}^{H \times W}\}_{v=1}^{V}$ are rendered from the 3D shape, \eg using the projection method in~\cite{pointclipv2_2023}. 
These $V$ multi-view images together with the text prompts $T$ of $R$ parts are passed to the VLM to get the knowledge in 2D spaces. 
For a B-VLM, a set of $\beta$ bounding boxes, $B^v=\{b^i\}_{i=1}^\beta$, is obtained from the $v$-th image, with $b^i \in \mathbb{R}^{4+R}$ encoding the box coordinates and the probabilities of the $R$ parts. 
For a P-VLM, a pixel-wise prediction map $S^v \in \mathbb{R}^{H \times W \times R}$ is acquired from the $v$-th image. 
We apply \textit{knowledge extraction} to each $B^v$ and each $S^v$ to obtain a  readily distillable knowledge $\mathcal{K}$ in the 3D space, as illustrated in Fig.~\ref{fig:main_fig}\bi{~(}b\bi{)} and Fig.~\ref{fig:main_fig}\bi{~(}c\bi{)}, respectively.

\bi{For a B-VLM, bounding boxes can directly be treated as the teacher knowledge.}
For a P-VLM, knowledge extraction starts by applying connected-component labeling~\cite{connectedcomp_2019} to $S^v$ to get a set of $\rho$ segmentation components, $\{s^i \in \mathbb{R}^{H \times W \times R}\}_{i=1}^{\rho}$, indicating if the $r$-th part is with the highest probability in each pixel.
We summarize the process when applying a VLM to a rendered image and the part text prompts as
\vspace{-6px}
\begin{equation}
    \text{VLM}(I^v,T)= \begin{cases}
    B^v=\{b^i\}_{i=1}^\beta, & \text{for B-VLM}, \\
    \mathbb{C}(S^v)=\{s^i\}_{i=1}^\rho, & \text{for P-VLM}, 
    \end{cases}
    \label{eq:vlm_out}
\end{equation}
where $\mathbb{C}$ denotes connected-component labeling. 

We then back-project each box $b^i$ or each prediction map $s^i$ to the 3D space, \ie
\vspace{-6px}
\begin{equation}
    k^i=(Y^i, M^i)= \begin{cases}
    \Gamma(b^i), & \text{for B-VLM}, \\
    \Gamma(s^i), & \text{for P-VLM}, 
    \end{cases}
    \label{eq:backproject}
\end{equation}
where $\Gamma$ denotes the back-projection operation with the camera parameters~\cite{camera_params_2000} used for multi-view image rendering, 
$Y^i \in \mathbb{R}^{N \times R}$ is the point-specific part probabilities, and $M^i \in \{0,1\}^{N}$ is the mask indicating which 3D points are covered by $b^i$ or $s^i$ in the 2D space.
The pair $(Y^i,M^i)$ yields a knowledge unit, $k^i$, upon which the knowledge re-scoring is performed in the backward distillation.

\bi{For} the 3D modality, a 3D encoder, \eg Point-M2AE~\cite{pointm2aezhang2023}, is applied to the point cloud and obtains per-point features, $O \in \mathbb{R}^{N \times E}$, capturing local and global geometrical information.
We then estimate point-wise part prediction, $\tilde{Y} \in {R}^{N \times R}$, by feeding the point features $O$ into the distillation head. The cross-modal distillation is performed by teaching the student network to align the part probability from the 3D modality $\tilde{Y}$ to their 2D counterparts $Y$ via minimizing our designated distillation loss.

\vspace{-10px}
\paragraph{Distillation loss.} 
Via Eq.~\ref{eq:vlm_out} and Eq.~\ref{eq:backproject}, we assume that $D$ knowledge units, $\mathcal{K}=\{k^d\}_{d=1}^{D}=\{Y^d,M^d\}_{d=1}^{D}$, are obtained from the multi-view images. 
The knowledge $\mathcal{K}$ exploits 2D appearance features and is incomplete as several 3D points are not covered by any 2D predictions, \ie issue $\boldsymbol{\mathcal{I}_1}$.
To distill this incomplete knowledge, we utilize a masked cross-entropy loss defined as
\begin{equation}
    \mathcal{L}_{distill} =-\sum_{d=1}^D \frac{1}{|M^d|}\sum_{n=1}^N \sum_{r=1}^R M^d_n C^d_n Z^d_{n,r} \text{log}(\tilde{Y}_{n,r}),
    \label{eq:loss_masked_crossent}
\end{equation}
where $C^d_n=\underset{r}{\text{max}}(Y^d_n(r))$ is the confidence score of $k^d$ on point $n$. 
$Z^d_{n,r}$ takes value $1$ if part $r$ receives the highest probability on $k^d$, and $0$ otherwise.
%
$|M^d|$ is the area covered by the mask $M^d$.

By minimizing Eq.~\ref{eq:loss_masked_crossent}, we teach the student network to align its prediction $\tilde{Y}$ to the distilled prediction $Y$ by considering the points covered by the mask and using the confidence scores as weights. 
%
Despite learning from incomplete knowledge, the student network extracts point features that capture geometrical information of the shape, thus enabling it to reasonably segment the points that are not covered by 2D predictions, hence addressing issue $\boldsymbol{\mathcal{I}_1}$. 
This can be regarded as interpolating the learned part probability in the feature spaces by the distillation head. 

As a distillation-based method, our method allows partial inconsistency among the extracted knowledge $\mathcal{K}=\{k^d\}_{d=1}^{D}$ caused by inaccurate VLM predictions, thereby alleviating issue $\boldsymbol{\mathcal{I}_2}$ of negative transfer.
In our method, the teacher network works on 2D appearance features, while the student network extracts 3D geometric features. 
After distillation via Eq.~\ref{eq:loss_masked_crossent}, the student network can exploit both appearance and geometric features from multiple shapes, hence mitigating issue $\boldsymbol{\mathcal{I}_3}$ of cross-modal transfer.
%
%
It is worth noting that unlike the conventional teacher-student models~\cite{distillation_hinton_2015,doublesim_distill_2021,3dv_distill_2021} which solely establish a one-to-one correspondence, we further \bi{re-score}
each knowledge unit $k^d$ based on its quality\bi{~(Sec.~\ref{subsec:knowledge_refinement})}, and improve distillation by suppressing low-quality knowledge units.



\subsection{\bi{Backward distillation: 3D to 2D}}
\label{subsec:knowledge_refinement}
\begin{figure}[t]
  \centering
  \includegraphics[keepaspectratio, width=0.95\columnwidth]{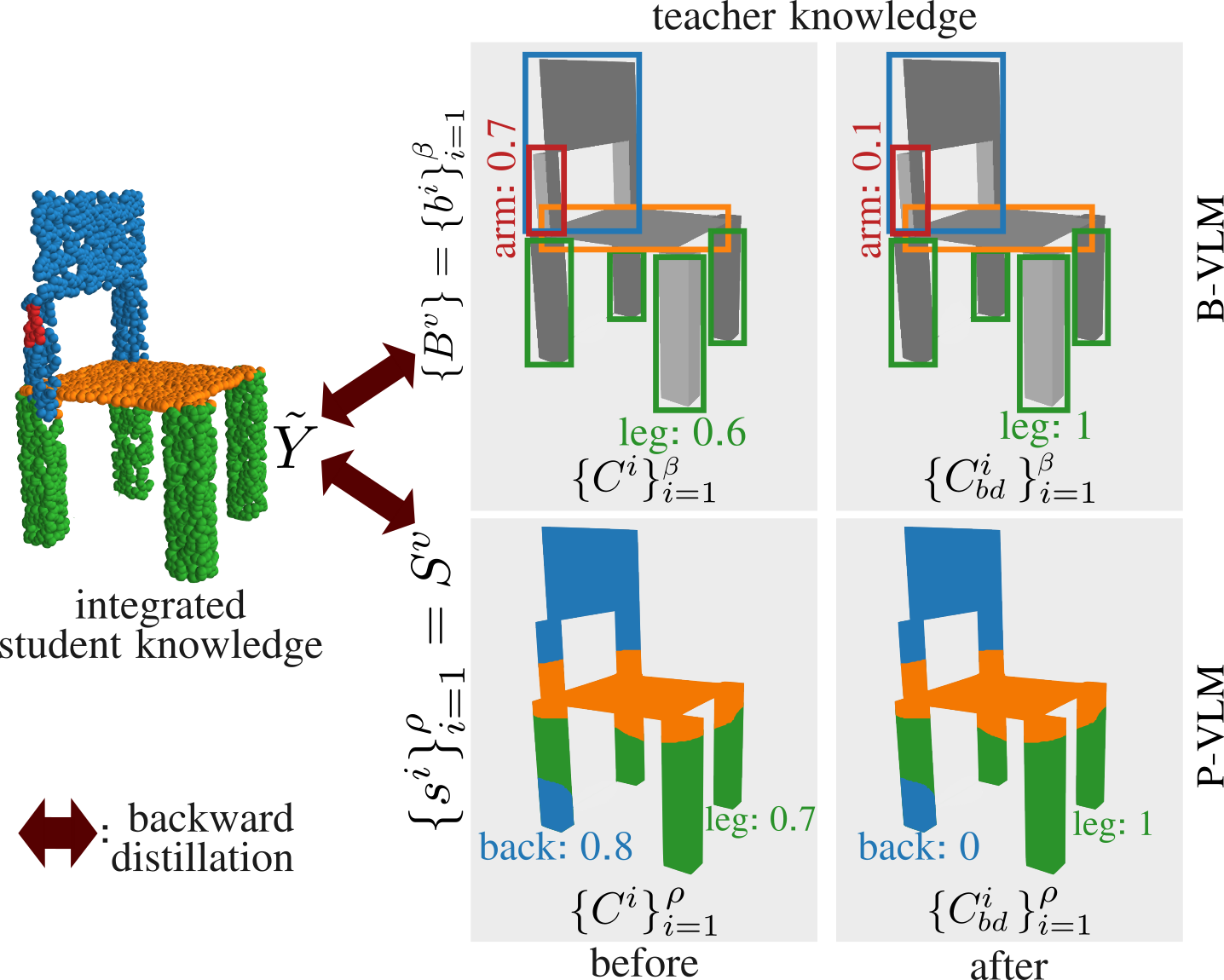}
  \caption{Given the VLM output of view $v$, $B^v$ or $S^v$, we display the confidence scores before ($C$) and after (\cam{$C_{bd}$}) performing backward distillation via Eq.~\ref{eq:knowledge_refinement}, with $Y$ and $M$  obtained via Eq.~\ref{eq:backproject}. 
  With backward distillation, inaccurate VLM predictions have lower scores, such as the arm box in B-VLM with the score reduced from 0.7 to 0.1, and vice versa.\vspace{-12px}} 
  \label{fig:knowledge_refinement}
\end{figure}

In Eq.~\ref{eq:loss_masked_crossent}, we consider all knowledge units $\{k^d\}_{d=1}^{D}$, weighted by their confidence scores. 
However, due to the potential VLM mispredictions, not all knowledge units are reliable. 
Hence, we refine the knowledge units by assigning higher scores to those of high quality and suppressing the low-quality ones. 
We observe that once the student network has thoroughly integrated the knowledge from the teacher, we can jointly refer both teacher and integrated student knowledge \bi{$\tilde{Y}'$} to achieve the goal, by re-scoring the confidence score $C^d$ to \cam{$C_{bd}^d$} as\bi{:}
\begin{equation}
    \cam{C_{bd}^d} = \frac{|M^d(\text{argmax}(Y^d) \cam{\Leftrightarrow} \text{argmax}(\tilde{Y}'))|}{|M^d|},
    \label{eq:knowledge_refinement}
\end{equation}
where \cam{$\Leftrightarrow$ denotes the element-wise equality (comparison) operation.}
In this way, each knowledge unit $k^d$ is \bi{re-scored:}
Those with high consensus between teacher $\mathcal{K}$ and integrated student knowledge \bi{$\tilde{Y}'$} have higher scores, such as those on the chair legs shown in Fig.~\ref{fig:knowledge_refinement}, and those with low consensus are suppressed by the reduced scores, such as those on the chair arm (B-VLM) and back (P-VLM) in Fig.~\ref{fig:knowledge_refinement}. 
Note that for simplicity, we only display two scores in each shape of Fig.~\ref{fig:knowledge_refinement} and show the average pixel-wise scores in P-VLM.
To justify that the student network has thoroughly integrated the teacher's knowledge, from initial knowledge $\tilde{Y}$ to integrated knowledge $\tilde{Y}'$, we track the moving average of the loss value for every epoch and see if the value in a subsequent epoch is lower than a specified threshold  $\tau$.
Afterward, the student network continues to learn with the 
\bi{re-scored knowledge $\mathcal{K}'$} by minimizing the loss in Eq.~\ref{eq:loss_masked_crossent} with $C$ being replaced by \cam{$C_{bd}$}\bi{, and produces the final part segmentation predictions $\tilde{Y}_f$}.

\subsection{\bi{Test-time alignment}}
\vspace{-3px}
In general, our method performs the alignment with a shape collection before the student network is utilized to carry the 3D shape part segmentation.
If such a pre-alignment is not preferred, we provide a special case of our method, test-time alignment (TTA), where the alignment is performed for every single shape in test time. To maintain the practicability, TTA needs to achieve a near-instantaneous completion.
To that end, TTA employs a readily used 3D encoder, \eg pre-trained Point-M2AE~\cite{pointm2aezhang2023}, freezes its weights, and only updates the learnable parameters in the distillation head, which significantly fastens the TTA completion. 

\subsection{Implementation Details}
\vspace{-5px}
The proposed framework is implemented in PyTorch~\cite{pytorch_2019} and is optimized for $25$ epochs via Adam optimizer~\cite{adam_2014} with a learning rate and batch size of $0.001$ and $16$, respectively.  
Unless further specified, the student network employs Point-M2AE~\cite{pointm2aezhang2023} pre-trained \cam{in a self-supervised way} on the ShapeNet55 dataset~\cite{shapenet55_2015} as the 3D extractor, freezes its weights, and only updates the learnable parameters in the distillation head. 
A multi-layer perceptron consisting of $4$ layers, with ReLU activation~\cite{relu_2018}, is adopted for the distillation head. 
To fairly compare with the competing methods~\cite{pointclipv1_2022,pointclipv2_2023,partslip_2023,satr_2023}, we follow their respective settings, including the used text prompts and the 2D rendering. 
Their methods render each shape into $10$ multi-view images, either from a sparse point cloud~\cite{pointclipv1_2022,pointclipv2_2023}, a dense point cloud~\cite{partslip_2023}, or a mesh shape~\cite{satr_2023}. 
\cam{Lastly, we follow~\cite{kelleher2020fundamentals,santry2023demystifying} to specify a small threshold value, $\tau=0.01$ in our backward distillation, and apply class-balance weighting~\cite{classbalance_2019} during the alignment, based on the VLM predictions in the zero-shot setting, with additional few-shot labels in the few-shot setting.}

\begin{table*}[!ht]
\centering
\begin{minipage}{\textwidth}
\caption{Zero-shot segmentation on the ShapeNetPart dataset, reported in mIoU (\%).*
\vspace{-6px}
}
\resizebox{0.99\textwidth}{!}{%
\begin{tabular}{cccccccccccccc}
\hline
VLM& Data type& Method& Airplane& Bag& Cap& Chair& Earphone& Guitar& Knife& Laptop& Mug& Table& Overall\\ \hline
\multirow{4}{*}{CLIP~\cite{clip_2021}} & \multirow{4}{*}{point cloud}& PointCLIP~\cite{pointclipv1_2022}       & \multicolumn{1}{c}{22.0} & \multicolumn{1}{c}{44.8} & \multicolumn{1}{c}{13.4} & \multicolumn{1}{c}{18.7} & \multicolumn{1}{c}{28.3} & \multicolumn{1}{c}{22.7} & \multicolumn{1}{c}{24.8} & \multicolumn{1}{c}{22.9} & \multicolumn{1}{c}{48.6} & \multicolumn{1}{c}{45.4} & \multicolumn{1}{c}{31.0} \\ 
                      &                              & PointCLIPv2~\cite{pointclipv2_2023}    & 35.7& 53.3& 53.1& 51.9& 48.1& 59.1& 66.7& 61.8& 45.5& 49.8& 48.4                \\
 & & \cam{OpenScene~\cite{openscene_peng2023}} & 34.4 & 63.8 & 56.1 & 59.8 & 62.6 & 69.3 & 70.1 & 65.4 & 51.0 &  60.4 & 52.9\\ 
                      &                              & Ours (TTA)& \textbf{37.5}& 62.6& 55.5& 56.4& 55.6& \textbf{71.7}& \textbf{76.9}& \textbf{67.4}& \textbf{53.5}& \textbf{62.9}& \textbf{53.8}                \\ 
                      &                              & Ours (Pre)& \textbf{40.6}& \textbf{75.6}& \textbf{67.2}& \textbf{65.0}& \textbf{66.3}& \textbf{85.8}& \textbf{79.8}& \textbf{92.6}& \textbf{83.1}& \textbf{68.7}& \textbf{63.9}                \\ \hline
\multirow{5}{*}{GLIP~\cite{glipv1_2022}} & \multirow{2}{*}{point cloud}& Ours (TTA)& 57.3& 62.7& 56.2& 74.2& 45.8& 60.6& 78.5& 85.7& 82.5& 62.9& 54.7                \\ 
                      &                              & Ours (Pre)& 69.3& 70.1& 67.9& 86.5& 51.2& 76.8& 85.7& 91.9& 85.6& 79.6& 64.1               \\ \cline{2-14} 
                      & \multirow{3}{*}{mesh}& SATR~\cite{satr_2023}            & 32.2& 32.1& 21.8& 25.2& 19.4& 37.7& 40.1& 50.4& 76.4& 22.4& 32.3                \\ 
                      &                              & Ours (TTA)& \textbf{53.2}& \textbf{61.8}& \textbf{44.9}& \textbf{66.4}& \textbf{43.0}& \textbf{50.7}& \textbf{66.3}& \textbf{68.3}& \textbf{83.9}& \textbf{58.8}& \textbf{49.5}                \\ 
                      &                              & Ours (Pre)& \textbf{64.8}& \textbf{64.4}& \textbf{51.0}& \textbf{67.4}& \textbf{48.3}& \textbf{64.8}& \textbf{70.0}& \textbf{83.1}& \textbf{86.5}& \textbf{79.3}& \textbf{56.3}         
                      \\ \hline
\end{tabular}
}
\label{tab:zeroshot_shapenet}
\small{*Results for other categories, including those of Table~\ref{tab:zeroshot_parte} and Table~\ref{tab:fewshot}, can be seen in the supplementary material.\vspace{-5px}}
\end{minipage}
\end{table*}

\begin{table*}[ht]
\centering
\caption{Zero-shot segmentation on the PartNetE dataset, reported in mIoU (\%). 
\vspace{-6px}
}
\resizebox{0.99\textwidth}{!}{%
\begin{tabular}{cccccccccccccc}
\hline
VLM                   & Data type                    & Method       & Bottle& Cart& Chair& Display& Kettle& Knife& Lamp& Oven& Suitcase& Table& Overall           \\ \hline
\multirow{2}{*}{GLIP~\cite{glipv1_2022}} & \multirow{2}{*}{point cloud} & PartSLIP~\cite{partslip_2023}    & 76.3& 87.7& 60.7& 43.8& 20.8& 46.8& 37.1& 33.0& 40.2& 47.7& 27.3\\ 
                      &                              & Ours (TTA) & \textbf{77.4}& \textbf{88.5}& \textbf{74.1}& \textbf{50.5}& \textbf{24.2}& \textbf{59.2}& \textbf{58.8}& \textbf{34.2}& \textbf{43.2}& \textbf{50.2}& \textbf{39.9}\\ \hline
\end{tabular}
}
\vspace{-10px}
\label{tab:zeroshot_parte}
\end{table*}
\vspace{-6px}
\section{Experiments}
\label{sec:experiments}
\vspace{-3px}
\subsection{Dataset and evaluation metric}
\label{dataset}
\vspace{-3px}
We evaluate the effectiveness of our method on two main benchmark datasets, ShapeNetPart~\cite{shapenet_2016} and PartNetE~\cite{partslip_2023}.
While ShapeNetPart dataset contains 16 categories with a total of 31,963 shapes, PartNetE
contains 2,266 shapes, covering 45 categories. 
The mean intersection over union (mIoU)~\cite{partnet_2019} is adopted to evaluate the segmentation results on the test-set data, measured against the ground truth label. 

\vspace{-2px}
\subsection{Zero-shot segmentation}
\label{subsec:zeroshot}
\vspace{-3px}
To compare with the competing methods~\cite{pointclipv1_2022,pointclipv2_2023,satr_2023,partslip_2023}, we adopt each of their settings and report their mIoU performances from their respective papers. Specifically, for P-VLM, we follow PointCLIP~\cite{pointclipv1_2022} and PointCLIPv2~\cite{pointclipv2_2023} to utilize CLIP~\cite{clip_2021} with ViT-B/32~\cite{vit_2020} backbone and use their pipeline to obtain the pixel-wise predictions from CLIP. For B-VLM, a GLIP-Large model~\cite{glipv1_2022} is employed in our method to compare with PartSLIP and SATR which also use the same model. While most competing methods report their performances on the ShapeNetPart dataset, PartSLIP evaluates its method
on the PartNetE dataset. \cam{In addition, we compare with OpenScene~\cite{openscene_peng2023} by extending it for 3D part segmentation and use the same Point-M2AE~\cite{pointm2aezhang2023} backbone and VLM CLIP for a fair comparison.}

Accordingly, we carry out the comparison separately to ensure fairness, based on the employed VLM model and the shape data type, \ie point cloud or mesh data, as shown in Tables~\ref{tab:zeroshot_shapenet} and~\ref{tab:zeroshot_parte}. In Table~\ref{tab:zeroshot_shapenet}, we provide two versions of our method, including test-time alignment (TTA) and pre-alignment (Pre) with a collection of shapes from the train-set data. Note that in the Pre version, our method does not use any labels (only unlabeled shape data are utilized).

First, we compare our method to PointCLIP and PointCLIPv2 (both utilize CLIP) on the zero-shot segmentation for the ShapeNetPart dataset, as can be seen in the first part of Table~\ref{tab:zeroshot_shapenet}. It is evident that our method for both TTA and pre-alignment versions achieves substantial improvements in all categories. For the overall mIoU, calculated by averaging the mIoUs from all categories, our method attains 5.4\% and 15.5\% higher mIoU for TTA and pre-alignment versions, respectively, compared to the best mIoU from the other methods. 
Such results reveal that our method which simultaneously exploits appearance and geometric features can better aggregate the 2D predictions for 3D part segmentation than directly averaging the corresponding 2D predictions as in the competing methods, where geometric evidence is not explored.
\cam{We further compare with OpenScene~\cite{openscene_peng2023} under the same setting as ours (Pre) and our method substantially outperforms it. One major reason is that our method can handle the inconsistency of VLM predictions (issue $\boldsymbol{\mathcal{I}}_2$) better by backward distillation.}

Next, as shown in the last three rows of Table~\ref{tab:zeroshot_shapenet}, we compare our method to SATR~\cite{satr_2023} that works on mesh data shapes. 
To obtain the mesh face predictions, we propagate the point predictions via a nearest neighbors approach as in~\cite{meshdata_shapenet_2017}, where each face is voted from its five nearest points.
Our method achieves 17.2\% and 24\% higher overall mIoU than SATR for TTA and pre-alignment versions, respectively.
Then, we compare our method with PartSLIP~\cite{partslip_2023} in Table~\ref{tab:zeroshot_parte} wherein only results from TTA are provided since the PartNetE dataset does not provide train-set data.
One can see that our method consistently obtains better segmentations, with 12.6\% higher overall mIoU than PartSLIP.

In PartSLIP and SATR, as GLIP is utilized, the uncovered 3D regions (issue $\boldsymbol{\mathcal{I}_1}$) could be intensified by possible undetected areas, and the negative transfer (issue $\boldsymbol{\mathcal{I}_2}$) may also be escalated due to semantic leaking, where the box predictions cover pixels from other semantics. On the other hand, our method can better alleviate these issues, thereby achieving substantially higher mIoU scores. In our method, the pre-alignment version achieves better segmentation results than TTA. This is expected since in the pre-alignment version, the student network can distill the knowledge from a collection of shapes, instead of individual shape. 

\begin{figure*}[t]
  \centering
  \includegraphics[keepaspectratio, width=0.95\textwidth]{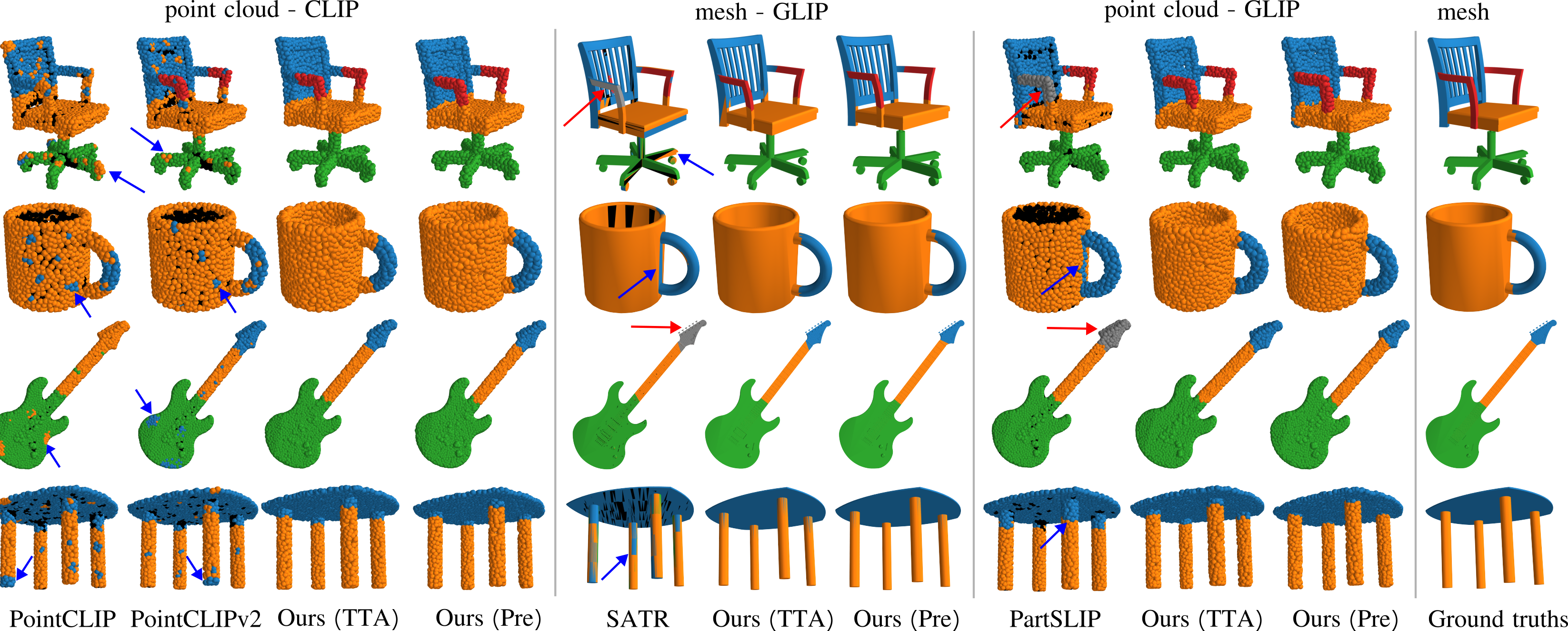}
  \vspace{-7px}
  \caption{Visualization of the zero-shot segmentation results, drawn in different colors, on the ShapeNetPart dataset. 
  We render PartSLIP results on the ShapeNetPart data to have the same visualization of shape inputs. While occluded and undetected regions (issue $\boldsymbol{\mathcal{I}_1}$) are shown with black and gray colors, respectively, the blue and red arrows highlight several cases of issues $\boldsymbol{\mathcal{I}_2}$ and $\boldsymbol{\mathcal{I}_3}.$ \vspace{-5px}}
  \label{fig:qualitative_result}
\end{figure*}
Besides foregoing quantitative comparisons, a qualitative comparison of the segmentation results is presented in Fig.~\ref{fig:qualitative_result}. It is readily observed that the competing methods suffer from the lack of 3D segmentation for the uncovered regions (issue $\boldsymbol{\mathcal{I}_1}$) caused by either occlusion or not being covered by any bounding box, drawn with black and gray colors, respectively. Moreover, these methods may also encounter negative transfers caused by inaccurate VLM outputs (issue $\boldsymbol{\mathcal{I}_2}$), such as those pointed by blue arrows, with notably degraded outcomes in SATR due to semantic leaking. Nonetheless, our method performs cross-modal distillation and alleviates these two issues, as can be seen in Fig.~\ref{fig:qualitative_result}. In addition, due to a direct transfer of 2D predictions to 3D space which relies on each independent shape as in the competing methods, erroneous 2D predictions will just remain as incorrect 3D segmentation (issue $\boldsymbol{\mathcal{I}_3}$), such as the missed detected chair arms and guitar heads pointed by red arrows. Our method also addresses this issue, by exploiting geometrical features across multiple shapes.

\begin{table*}[ht]
\centering
\caption{Few-shot segmentation on the PartNetE dataset, reported in mIoU (\%). 
\vspace{-7px}
}
\resizebox{0.95\textwidth}{!}{%
\begin{tabular}{ccccccccccccc}
\hline
\multicolumn{2}{c}{Method}                                         & \multicolumn{1}{c}{Bottle} & \multicolumn{1}{c}{Cart} & \multicolumn{1}{c}{Chair} & \multicolumn{1}{c}{Display} & \multicolumn{1}{c}{Kettle} & \multicolumn{1}{c}{Knife} & \multicolumn{1}{c}{Lamp} & \multicolumn{1}{c}{Oven} & \multicolumn{1}{c}{Suitcase} & \multicolumn{1}{c}{Table} & \multicolumn{1}{c}{Overall} \\ \hline
\multirow{5}{*}{Non-VLM-based}& PointNet++~\cite{pointnet++_2017} &                           27.0&                                 11.6&                          42.2&                            30.2&                         28.6&                              22.2&                           10.5&                            19.4&                               3.3&                         7.3&                          20.4\\ 
 & PointNext~\cite{pointnext_2022}  & 67.6& 47.7& 65.1& 53.7& 60.6& 59.7& 55.4& 36.8& 14.5&                         22.1&                          40.6\\ 
 & ACD~\cite{acd_2020} &                           22.4&                                 31.5&                          39.0&                            29.2&                         40.2&                              39.6&                           13.7&                            8.9&                               13.2&                         13.5&                          23.2\\ 
 & Prototype~\cite{prototype_2021} &                           60.1&                                 36.8&                          70.8&                            67.3&                         62.7&                              50.4&                           38.2&                            36.5&                               35.5&                         25.7&                          44.3\\
 & \cam{Point-M2AE~\cite{pointm2aezhang2023}} & 72.4 & 74.5 & 83.4 & 74.3 & 64.3 & 68.0 & 57.6& 53.3& 57.5& 33.6& 56.4 \\ \hline
\multirow{2}{*}{\makecell{VLM-based\\  
 (GLIP~\cite{glipv1_2022})}}  & PartSLIP~\cite{partslip_2023} &                           83.4&                                 88.1&                          85.3&                            84.8&                         77.0&                              65.2&                           60.0&                            \textbf{73.5}&                               70.4&                         42.4&                          59.4\\  
& Ours &                           \textbf{84.6}&                                 \textbf{90.1}&                          \textbf{88.4}&                            \textbf{87.4}&                         \textbf{78.6}&                              \textbf{71.4}&                           \textbf{69.2}&                            72.8&                               \textbf{73.4}&                         \textbf{63.3}&                          \textbf{65.9}\\ \hline
\end{tabular}
}
\vspace{-12px}
\label{tab:fewshot}
\end{table*}

\vspace{-4px}
\subsection{Few-shot segmentation}
\label{subsec:fewshot}
\vspace{-4px}
We further demonstrate the effectiveness of our method in a few-shot scenario by following the setting used in PartSLIP~\cite{partslip_2023}. Specifically, we employ the fine-tuned GLIP model~\cite{glipv1_2022} provided by PartSLIP via 8-shot labeled shapes of the PartNetE dataset~\cite{shapenet_2016} for each category. In addition to the alignment via Eq.~\ref{eq:loss_masked_crossent}, we ask the student network to learn parameters that minimize both Eq.~\ref{eq:loss_masked_crossent} and a standard cross-entropy loss for segmentation on the 8 labeled shapes.

As shown in Table~\ref{tab:fewshot}, the methods dedicated to few-shot 3D segmentation, ACD~\cite{acd_2020} and Prototype~\cite{prototype_2021}, are adapted to PointNet++~\cite{prototype_2021} and PointNext~\cite{pointnext_2022} backbones, respectively, and can improve the performances (on average) of these backbones. PartSLIP, on the other hand, leverages multi-view GLIP predictions for 3D segmentation and further improves the mIoU, but there are still substantial performance gaps compared to our method which distills the GLIP predictions instead. \cam{We also present the results from fine-tuning Point-M2AE with the few-shot labels, which shows lower performances than ours, highlighting the significant contribution of our distillation framework. For more qualitative results, see the supplementary materials.}

\subsection{Leveraging generated data}
\vspace{-5pt}
\label{subsec:generateddata}
Since only unlabeled 3D shape data are required for our method to perform cross-modal distillation, existing generative models~\cite{dit3d_2023,shapeaspoint_2021} can facilitate an effortless generation of 3D shapes, and the generated data can be smoothly incorporated into our method. Specifically, we first adopt DiT-3D~\cite{dit3d_2023} which is pre-trained on the ShapeNet55 dataset~\cite{shapenet55_2015} to generate point clouds of shapes, 500 shapes for each category, and further employ SAP~\cite{shapeaspoint_2021} to transform the generated point clouds into mesh shapes. These generated mesh shapes can then be utilized in our method for distillation. Table~\ref{tab:generated_data_result} shows the results evaluated on the test-set data of ShapeNetPart~\cite{shapenet_2016} and COSEG~\cite{coseg_2012} datasets for several shape categories, using GLIP as the VLM.

One can see that with distilling from the generated alone, our method already achieves competitive results on the ShapeNetPart dataset compared to distilling from the train-set data. Since the generated data via DiT-3D is pre-trained on the ShapeNet55 dataset which contains the ShapeNetPart data, we also evaluate its performance on the COSEG dataset to show that such results can be well transferred to shapes from another dataset. Finally, Table~\ref{tab:generated_data_result} (the last row) reveals that using generated data as a supplementary knowledge source can further increase the mIoU performance. 
\cam{Such results suggest that if a collection of shapes is available, generated data can be employed as supplementary knowledge sources, which can improve the performance. On the other hand, if a collection of shapes does not exist, generative models can be employed for shape creation and subsequently used in our method as the knowledge source.}

\vspace{-3pt}
\subsection{Ablation studies}
\label{ablation_analysis}
\vspace{-5px}
\paragraph{Proposed components.} We perform ablation studies on the proposed components, and the mIoU scores in 2D\footnote{Calculated between the VLM predictions and their corresponding 2D ground truths projected from 3D, and weighted by the confidence scores. See supplementary material for the details.} and 3D spaces on three categories of the ShapeNetPart dataset are shown in (1) to (9) of Table~\ref{tab:ablation}. In (1), only GLIP box predictions are utilized to get 3D segmentations, \ie part labels are assigned by voting from all visible points within the multi-view box predictions. These numbers serve as baselines and are subject to issues $\boldsymbol{\mathcal{I}_1}$$\sim$$\boldsymbol{\mathcal{I}_3}$. 
In (2) and (3), 3D segmentations are achieved via \emph{forward distillation} from the GLIP predictions to the student network using Eq.~\ref{eq:loss_masked_crossent}, for test-time alignment (TTA) and pre-alignment (Pre) versions, resulting in significant improvements compared to the baselines, with more than 10\% and 14\% higher mIoUs, respectively. Such results demonstrate that the proposed cross-modal distillation can better utilize the 2D multi-view predictions for 3D part segmentation, alleviating $\boldsymbol{\mathcal{I}_1}$$\sim$$\boldsymbol{\mathcal{I}_3}$.

\begin{table}[t]
\centering
\caption{Segmentation mIoU (\%) by leveraging generated data.\vspace{-5px}
}
\resizebox{0.95\columnwidth}{!}{%
\begin{tabular}{cccccc}
\hline
\multirow{2}{*}{Distilled data} & \multicolumn{3}{c}{ShapeNetPart~\cite{shapenet_2016}} & \multicolumn{2}{c}{COSEG~\cite{coseg_2012}} \\ \cline{2-6} 
                                & Airplane  & Chair  & Guitar  & Chair       & Guitar      \\ \hline
Train-set (baseline)                     &           69.3&        86.2&         76.8&             96.4&             68.0\\ 
Gen. data                  &           69.0&        85.3&         75.6&             96.1&             67.5\\ 
Gen. data \& train-set          &           70.8&        88.4&         78.3&             97.4&             70.2\\ \hline
\end{tabular}
}
\label{tab:generated_data_result}
\vspace{-5px}
\end{table}
\begin{table}[t]
\centering
\caption{Ablation study on the proposed method. \vspace{-5px}
}
\resizebox{1\columnwidth}{!}{
\begin{tabular}{cccccllllll}
\hline
\multirow{2}{*}{No} & \multirow{2}{*}{VLM}  & \multirow{2}{*}{Pre}& \multirow{2}{*}{BD}& \multirow{2}{*}{\makecell{Student\\network}}& \multicolumn{2}{c}{Airplane}                        & \multicolumn{2}{c}{Chair}                           & \multicolumn{2}{c}{Knife}                          \\ 
                       &                       &                     &                      &                                 & \multicolumn{1}{c}{2D}   & \multicolumn{1}{c}{3D}   & \multicolumn{1}{c}{2D}   & \multicolumn{1}{c}{3D}   & \multicolumn{1}{c}{2D}   & \multicolumn{1}{c}{3D}   \\ \hline
(1)                      & \multirow{9}{*}{\makecell{GLIP\\\cite{glipv1_2022}}} & & & & \multicolumn{1}{c}{42.8} & \multicolumn{1}{c}{40.2} & \multicolumn{1}{c}{60.2} & \multicolumn{1}{c}{60.1} & \multicolumn{1}{c}{53.6} & \multicolumn{1}{c}{57.2} \\ 
(2)                      &                       & & & \ding{51}&                          42.8&                          56.2&                          60.2&                          73.5&                          53.6&                          77.6\\ 
(3)                      &                       & \ding{51}& & \ding{51}&                          42.8&                          64.3&                          60.2&                          84.2&                          53.6&                          84.5\\ 
(4)&                       & & \ding{51}& \ding{51}&                          44.3&                          57.3&                          61.7&                          74.2&                          54.8&                          78.5\\ 
                       (5)&                       & \ding{51}& \ding{51}& \ding{51}&                          48.2&                          \textbf{69.3}&                          63.2&                          86.5&                          55.0&                          \textbf{85.7}\\ 
                       (6)&                       & \ding{51}& \ding{51}& exclude $\boldsymbol{\mathcal{I}_1}$&                          48.2&                          62.5&                          63.2&                          80.4&                          55.0&                          81.2\\ 
                       (7)&                       & \ding{51}& \ding{51}& w/o pretrain&                          48.2&                          69.1&                          63.2&                          \textbf{86.7}&                          55.0&                          85.3\\ \hline
(8)                     & \multirow{2}{*}{\makecell{CLIP\\\cite{clip_2021}}} & \ding{51}& & \ding{51}&                          34.6&                          38.4&                          50.4&                          63.6&                          66.8&                          77.4\\ 
(9)                     &                       & \ding{51}& \ding{51}& \ding{51}&                          37.8&                          \textbf{40.6}&                          54.2&                          \textbf{65.0}&                          68.4&                          \textbf{78.9}\\ \hline
\end{tabular}}
\label{tab:ablation}
\vspace{-9px}
\end{table}

We further add \emph{backward distillation} (BD) in (4) and (5), which substantially improves the knowledge source in 2D, \eg from 42.8\% to 48.2\% for the airplane category in the pre-alignment version, and subsequently enhances the 3D segmentation. We observe a higher impact (improvement) on the pre-alignment compared to TTA versions, \ie in (4) and (5), as the student network of the former can better integrate the knowledge from a collection of shapes. A similar trend of improvement can be observed for a similar ablation performed with CLIP~\cite{clip_2021} used as the VLM (in (8) and (9)).

In (6), we exclude our method's predictions for those uncovered points to simulate issue $\boldsymbol{\mathcal{I}_1}$, and the reduced mIoUs compared to (5), \eg from 86.5\% to 80.4\% for the chair category, reveal that our method can effectively alleviate issue $\boldsymbol{\mathcal{I}_1}$. Finally, instead of using pre-trained weights of Point-M2AE~\cite{pointm2aezhang2023} and freezing them as the 3D decoder as in (5), we initialize these weights (by default PyTorch~\cite{pytorch_2019} initialization) and set them to be learnable as in (7). Both settings produce comparable results (within 0.4\%). The main purpose of using the pre-trained weights and freezing them is for faster convergence, especially for the test-time alignment purpose. Please refer to the supplementary material for the comparison of convergence curves. 

\begin{figure}[t]
  \centering
  \includegraphics[keepaspectratio, width=1\columnwidth]{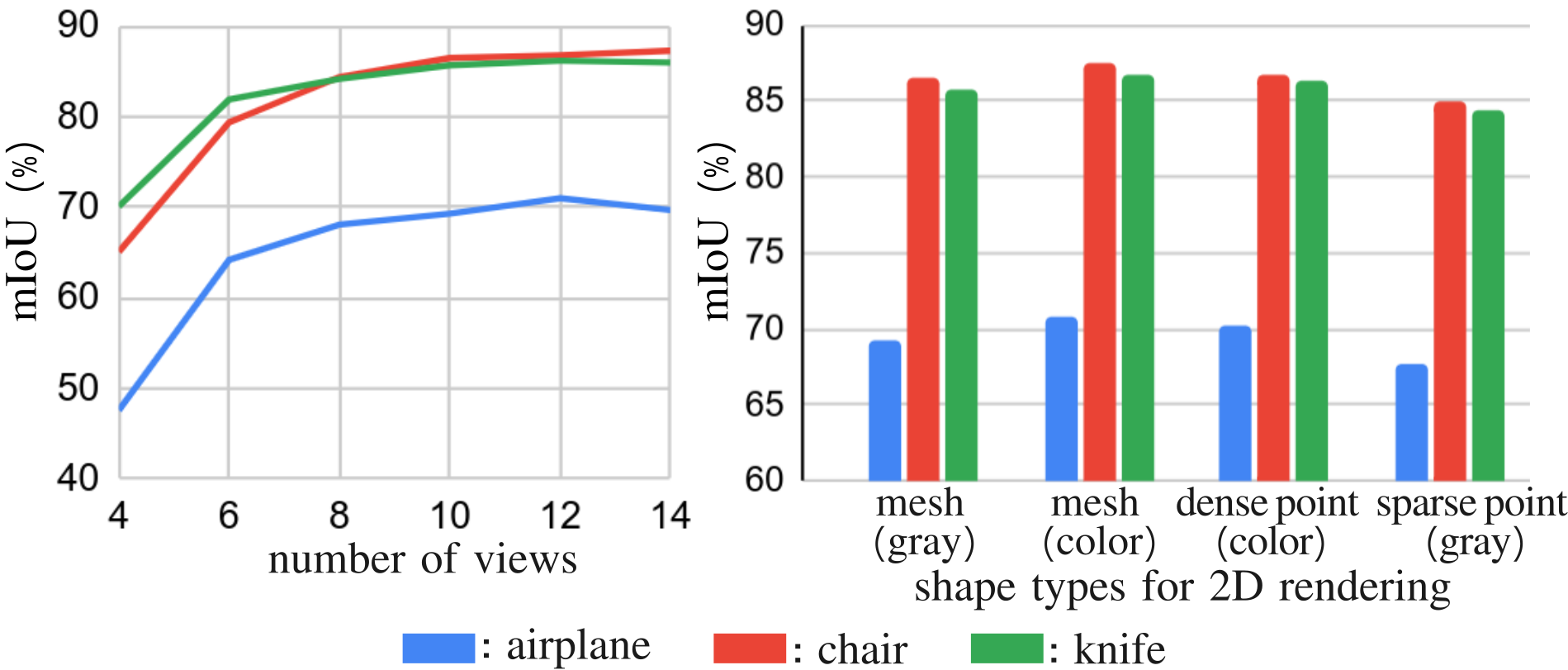}
  \vspace{-12px}
  \caption{Ablation study on number of views and various shape types for 2D multiview rendering on the ShapeNetPart dataset. 
  \vspace{-10px}}
  \label{fig:ablation_view_shapetype}
\end{figure}

\vspace{-15px}
\paragraph{Number of views.} We render $V=10$ multi-view images for each shape input in our main experiment, and Fig.~\ref{fig:ablation_view_shapetype} (left) shows the mIoU scores with different values of $V$.  A substantial drop is observed when utilizing $V<6$, and small increases are obtained when a larger $V$ is used.

\vspace{-15px}
\paragraph{Various shape types for 2D multi-view rendering.} We render 10 multi-view images from various shape data types, \ie (i) gray mesh, (ii) colored mesh, (iii) dense colored point cloud ($\sim$300k points) as used in PartSLIP~\cite{partslip_2023}, and (iv) sparse gray point cloud (2,048 points) using PyTroch3D~\cite{pytorch3d_2020} and the rendering method in~\cite{pointclipv2_2023} to render (i)-(iii) and (iv), respectively. Fig.~\ref{fig:ablation_view_shapetype} (right) summarizes such results on the ShapeNetPart dataset, with GLIP used as the VLM. Note that the first three shape types produce comparable mIoUs with slightly higher scores when colored mesh or dense colored point cloud is utilized. When sparse gray point cloud data type is used, a mild mIoU decrease is observed. Please refer to the supplementary material to see more results for (i)-(iv). 

\vspace{-10px}
\paragraph{Limitation.} The main limitation of our method is that the segmentation results are impacted by the quality of the VLM predictions, where VLMs are generally pre-trained to recognize object- or sample-level categories (not part-level of object categories). For instance, GLIP can satisfactorily locate part semantics for the chair category but with lower qualities for the earphone category, while CLIP can favorably locate part semantics for the earphone category but with less favorable results for the airplane category. Hence, exploiting multiple VLMs can be a potential future work. Nonetheless, the proposed method which currently employs a single VLM model can already boost the segmentation results significantly compared to the existing methods.   

\vspace{-2px}
\section{Conclusion}
\label{sec:conclusion}
\vspace{-1px}
We present a cross-modal distillation framework that transfers 2D knowledge from a vision-language model (VLM) to facilitate 3D shape part segmentation, which generalizes well to both VLM with bounding-box and pixel-wise predictions. In the proposed method, backward distillation is introduced to enhance the quality of 2D predictions and subsequently improve the 3D segmentation. The proposed approach can also leverage existing generative models for shape creation and can be smoothly incorporated into the method for distillation. With extensive experiments, the proposed method is compared with existing methods on widely used benchmark datasets, including ShapeNetPart and PartNetE, and consistently outperforms existing methods with substantial margins both in zero-shot and few-shot scenarios on 3D data in point
clouds or mesh shapes.
\vspace{-0.1in}
\paragraph{Acknowledgment.}
This work was supported in part by the National Science and Technology Council (NSTC) under grants 112-2221-E-A49-090-MY3, 111-2628-E-A49-025-MY3, 112-2634-F-006-002 and 112-2634-F-A49-007. This work was funded in part by MediaTek and NVIDIA.
\maketitlesupplementary

\begin{table*}[t]
\centering
\caption{Zero-shot segmentation on all 16 categories of the ShapeNetPart dataset~\cite{shapenet_2016}, reported mIoU (\%). In this table, TTA and Pre denote the test-time alignment and pre-alignment versions of our method, while VLM stands for vision-language model (see main paper for details).}
\resizebox{0.99\textwidth}{!}{
\begin{tabular}{c|cccc|cc|ccc}
\toprule
\multirow{3}{*}{Category} & \multicolumn{4}{c|}{VLM - CLIP~\cite{clip_2021}}                           & \multicolumn{5}{c}{VLM - GLIP~\cite{glipv1_2022}}                                               \\ \cmidrule{6-10} 
                          & \multicolumn{4}{c|}{point cloud input}                    & \multicolumn{2}{c|}{point cloud input} & \multicolumn{3}{c}{mesh input}       \\ \cmidrule{2-10} 
                          & PointCLIP~\cite{pointclipv1_2022} & PointCLIP v2~\cite{pointclipv2_2023} & Ours (TTA)    & Ours (Pre)    & Ours (TTA)        & Ours (Pre)        & SATR~\cite{satr_2023} & Ours (TTA)    & Ours (Pre)    \\ \midrule
Airplane                  & 22.0      & 35.7         & \textbf{37.5} & \textbf{40.6} & 57.3              & 69.3              & 32.2 & \textbf{53.2} & \textbf{64.8} \\ 
Bag                       & 44.8      & 53.3         & \textbf{62.6} & \textbf{75.6} & 62.7              & 70.1              & 32.1 & \textbf{61.8} & \textbf{64.4} \\ 
Cap                       & 13.4      & 53.1         & \textbf{55.5} & \textbf{67.2} & 56.2              & 67.9              & 21.8 & \textbf{44.9} & \textbf{51.0} \\ 
Car                       & 30.4      & 34.5         & \textbf{36.4} & \textbf{41.2} & 32.4              & 39.2              & 22.3 & \textbf{30.2} & \textbf{32.3} \\ 
Chair                     & 18.7      & 51.9         & \textbf{56.4} & \textbf{65.0} & 74.2              & 86.5              & 25.2 & \textbf{66.4} & \textbf{67.4} \\ 
Earphone                  & 28.3      & 48.1         & \textbf{55.6} & \textbf{66.3} & 45.8              & 51.2              & 19.4 & \textbf{43.0} & \textbf{48.3} \\ 
Guitar                    & 22.7      & 59.1         & \textbf{71.7} & \textbf{85.8} & 60.6              & 76.8              & 37.7 & \textbf{50.7} & \textbf{64.8} \\ 
Knife                     & 24.8      & 66.7         & \textbf{76.9} & \textbf{79.8} & 78.5              & 85.7              & 40.1 & \textbf{66.3} & \textbf{70.0} \\ 
Lamp                      & 39.6      & 44.7         & \textbf{45.8} & \textbf{63.1} & 34.5              & 43.5              & 21.6 & \textbf{30.5} & \textbf{35.2} \\ 
Laptop                    & 22.9      & 61.8         & \textbf{67.4} & \textbf{92.6} & 85.7              & 91.9              & 50.4 & \textbf{68.3} & \textbf{83.1} \\ 
Motorbike                 & 26.3      & 31.4         & \textbf{33.4} & \textbf{38.2} & 30.6              & 37.8              & 25.4 & \textbf{28.8} & \textbf{32.5} \\ 
Mug                       & 48.6      & 45.5         & \textbf{53.5} & \textbf{83.1} & 82.5              & 85.6              & 76.4 & \textbf{83.9} & \textbf{86.5} \\ 
Pistol                    & 42.6      & 46.1         & \textbf{48.2} & \textbf{55.8} & 39.6              & 48.5              & 34.1 & \textbf{37.4} & \textbf{40.9} \\ 
Rocket                    & 22.7      & 46.7         & \textbf{49.3} & \textbf{49.5} & 36.8              & 48.9              & 33.2 & \textbf{41.1} & \textbf{45.3} \\ 
Skateboard                & 42.7      & 45.8         & \textbf{47.7} & \textbf{49.2} & 34.2              & 43.5              & 22.3 & \textbf{26.2} & \textbf{34.5} \\ 
Table                     & 45.4      & 49.8         & \textbf{62.9} & \textbf{68.7} & 62.9              & 79.6              & 22.4 & \textbf{58.8} & \textbf{79.3} \\ 
Overall                   & 31.0      & 48.4         & \textbf{53.8} & \textbf{63.9} & 54.7              & 64.1              & 32.3 & \textbf{49.5} & \textbf{56.3} \\ \bottomrule
\end{tabular}
}
\label{tab:shapenet_miou}
\end{table*}
\begin{table}[t]
\centering
\caption{Segmentation on all 45 categories of the PartE dataset~\cite{partslip_2023}, reported in mIoU (\%). In this table, TTA denotes our method with test-time alignment (see main paper for details).}
\resizebox{0.99\columnwidth}{!}{
\begin{tabular}{c|cc|cc}
\toprule
\multirow{2}{*}{Category} & \multicolumn{2}{c|}{Zero-shot} & \multicolumn{2}{c}{Few-shot}  \\ \cmidrule{2-3} \cmidrule{4-5}
                          & PartSLIP~\cite{partslip_2023}      & Ours (TTA)    & PartSLIP~\cite{partslip_2023}      & Ours
                          \\ \midrule
Bottle                    & 76.3          & \textbf{77.4} & 83.4          & \textbf{84.6} \\
Box                       & 57.5          & \textbf{69.7} & 84.5          & \textbf{87.9} \\
Bucket                    & 2.0           & \textbf{16.8} & 36.5          & \textbf{50.7} \\
Camera                    & 21.4          & \textbf{29.4} & 58.3          & \textbf{60.1} \\
Cart                      & 87.7          & \textbf{88.5} & 88.1          & \textbf{90.1} \\
Chair                     & 60.7          & \textbf{74.1} & 85.3          & \textbf{88.4} \\
Clock                     & \textbf{26.7} & 23.6          & \textbf{37.6} & 37.2          \\
Coffee machine            & 25.4          & \textbf{26.8} & 37.8          & \textbf{40.2} \\
Dishwasher                & 10.3          & \textbf{18.6} & 62.5          & \textbf{60.2} \\
Dispenser                 & 16.5          & \textbf{11.4} & 73.8          & \textbf{74.7} \\
Display                   & 43.8          & \textbf{50.5} & 84.8          & \textbf{87.4} \\
Door                      & 2.7           & \textbf{41.1} & 40.8          & \textbf{55.5} \\
Eyeglasses                & 1.8           & \textbf{59.7} & 88.3          & \textbf{91.1} \\
Faucet                    & 6.8           & \textbf{33.3} & 71.4          & \textbf{73.5} \\
Folding chair             & 91.7          & \textbf{89.7} & 86.3          & \textbf{90.7} \\
Globe                     & 34.8          & \textbf{90.0} & 95.7          & \textbf{97.4} \\
Kettle                    & 20.8          & \textbf{24.2} & 77.0          & \textbf{78.6} \\
Keyboard                  & 37.3          & \textbf{38.5} & 53.6          & \textbf{70.8} \\
Kitchenpot                & 4.7           & \textbf{36.8} & 69.6          & \textbf{69.7} \\
Knife                     & 46.8          & \textbf{59.2} & 65.2          & \textbf{71.4} \\
Lamp                      & 37.1          & \textbf{58.8} & 66.1          & \textbf{69.2} \\
Laptop                    & 27.0          & \textbf{37.1} & 29.7          & \textbf{40.0} \\
Lighter                   & 35.4          & \textbf{37.3} & 64.7          & \textbf{64.9} \\
Microwave                 & 16.6          & \textbf{23.2} & 42.7          & \textbf{43.8} \\
Mouse                     & \textbf{27.0} & 18.6          & 44.0          & \textbf{46.9} \\
Oven                      & 33.0          & \textbf{34.2} & \textbf{73.5} & 72.8          \\
Pen                       & 14.6          & \textbf{15.7} & 71.5          & \textbf{74.4} \\
Phone                     & 36.1          & \textbf{37.3} & 48.4          & \textbf{50.8} \\
Pliers                    & 5.4           & \textbf{51.9} & 33.2          & \textbf{90.4} \\
Printer                   & 0.8           & \textbf{3.3}  & 4.3           & \textbf{6.3}  \\
Refrigerator              & 20.2          & \textbf{25.2} & 55.8          & \textbf{58.1} \\
Remote                    & 11.5          & \textbf{13.2} & 38.3          & \textbf{40.7} \\
Safe                      & \textbf{22.4} & 18.2          & 32.2          & \textbf{58.6} \\
Scissors                  & 21.8          & \textbf{64.4} & 60.3          & \textbf{68.8} \\
Stapler                   & 20.9          & \textbf{65.1} & 84.8          & \textbf{86.3} \\
Storage furniture         & 29.5          & \textbf{30.6} & 53.6          & \textbf{56.5} \\
Suitcase                  & 40.2          & \textbf{43.2} & 70.4          & \textbf{73.4} \\
Switch                    & 9.5           & \textbf{30.3} & 59.4          & \textbf{60.7} \\
Table                     & 47.7          & \textbf{50.2} & 42.5          & \textbf{63.3} \\
Toaster                   & \textbf{13.8} & 11.4          & \textbf{60.0} & 58.7          \\
Toilet                    & 20.6          & \textbf{22.5} & 53.8          & \textbf{55.0} \\
Trash can                 & 30.1          & \textbf{49.3} & 22.3          & \textbf{70.0} \\
Usb                       & 10.9          & \textbf{39.1} & 54.4          & \textbf{64.3} \\
Washing machine           & 12.5          & \textbf{12.9} & 53.5          & \textbf{55.1} \\
Window                    & 5.2           & \textbf{45.3} & 75.4          & \textbf{78.1} \\
Overall                   & 27.3          & \textbf{39.9} & 59.4          & \textbf{65.9} \\ \bottomrule
\end{tabular}
}
\label{tab:parte_quan}
\end{table}

\section{3D segmentation scores for full categories}
We provide 3D segmentation scores, reported in mIoU, for full categories of the ShapeNetPart~\cite{shapenet_2016} and PartE~\cite{partslip_2023} datasets in Tables~\ref{tab:shapenet_miou} and~\ref{tab:parte_quan}, respectively. Table~\ref{tab:shapenet_miou} is associated with Table~1 in the main paper, while Table~\ref{tab:parte_quan} is associated with Tables~2 and 3. In Table~\ref{tab:shapenet_miou}, 16 categories of the ShapeNetPart dataset are reported, while 45 categories of the PartE dataset are presented in Table~\ref{tab:parte_quan}. For the tables, it is readily observed that the proposed method, \textbf{\emph{PartDistill}}, attains substantial improvements compared to the competing methods~\cite{pointclipv1_2022,pointclipv2_2023,satr_2023,partslip_2023} in most categories. 

\section{Evaluating 2D predictions}
In the ablation studies of our method's components presented in Table 5, we provide mIoU scores in 2D space, $\text{mIoU}^{\text{2D}}$, to evaluate the quality of the 2D predictions measured against the 2D ground truths before and after performing \emph{backward distillation} which re-scores the confidence scores of each knowledge unit. Here, the 2D ground truths are obtained by projecting the 3D mesh (faces) part segmentation labels to 2D space using the camera parameters utilized when performing 2D multi-view rendering. 

We first explain how to calculate the mIoU$^{\text{2D}}$ if a vision-language model (VLM) which outputs pixel-wise predictions (P-VLM) is used in our method and later explain if a VLM which outputs bounding-box predictions (B-VLM) is employed.  In each view, let $\{s^i\}_{i}^{\rho}$ be the prediction maps (see Eq.~1 in the main paper) of P-VLM with $C^i$ denoting the confidence score of $s^i$ and  $\mathcal{G}$ be the corresponding 2D ground truth. We first calculate the IoU$^{\text{2D}}$ for each semantic part $r$ as,
\begin{equation}
    \text{IoU}^{\text{2D}}(r)=\frac{\mathcal{I}(r)}{\mathcal{I}(r)+\lambda(r)+\gamma(r)}
    \label{eq:iou}
\end{equation}
where 
\begin{equation}
    \mathcal{I}(r)=\sum_{i \in \phi(r)}\text{Avg}(C^i)(s^{i} \cap \mathcal{G}^{r})
    \label{eq:firstterm},
\end{equation}

\begin{equation}
    \lambda(r)=\text{Avg}(C^{\phi(r)}) \left (\left (\bigcup_{i \in \phi(r)}s^{i} \right) \notin \mathcal{G}^{r} \right)
    \label{eq:secondterm},
\end{equation}

and

\begin{equation}
    \gamma(r)=  \mathcal{G}^{r} \notin  \bigcup_{i \in \phi(r)}s^{i}
    \label{eq:thirdterm},
\end{equation}
with $\phi(r)$ denoting a function returning indices of $\{s^i\}_{i=1}^{\rho}$ that predict part $r$, ``Avg'' denoting an averaging operation and $\mathcal{G}^{r}$ indicating the ground truth of part $r$.

While Eq.~\ref{eq:firstterm} represents the intersection of the pixels between the 2D predictions and the corresponding ground truths, weighted by their confidence scores, Eq.~\ref{eq:secondterm} tells the union of the 2D predictions pixels that do not intersect with the corresponding ground truths, which is weighted by the average of all confidence scores associated with part $r$. As for Eq.~\ref{eq:thirdterm}, it tells the ground truth pixels that do not intersect with the corresponding 2D predictions union. We then calculate the IoU$^{\text{2D}}$ score for each semantic part $r$ in every $v$ view and compute the mean of them as mIoU$^{\text{2D}}$.

Note that we involve the confidence scores as weights to calculate the mIoU$^{\text{2D}}$. This allows us to compare the quality of the 2D predictions before and after applying \emph{backward distillation}, using the confidence scores before and after this process. To compute the mIoU$^{\text{2D}}$ scores when a B-VLM is used in our method, we can use Eq.~\ref{eq:iou} with $s^i$ in Eq.~\ref{eq:firstterm}$\sim$~Eq.~\ref{eq:thirdterm} being replaced by $\mathcal{F}(b^i)$, where $\mathcal{F}$ denotes an operation excluding the background pixels covered in $b^i$.

\section{Additional visualizations}
\subsection{Visualization of few-shot segmentation}
\begin{figure*}[t]
  \centering
  \vspace{10pt}
  \includegraphics[keepaspectratio, width=0.98\textwidth]{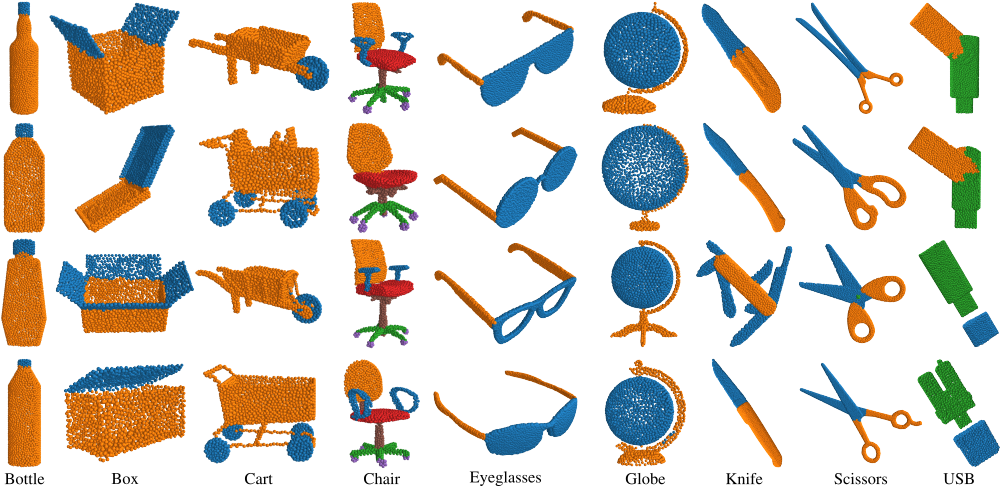}
  \caption{Visualization of few-shot segmentation results derived using our method on the PartE dataset~\cite{partslip_2023}. Each semantic part is drawn in different colors.\vspace{0px}}
  \label{fig:fewshot}
\end{figure*}
In Figure~\ref{fig:fewshot}, we present several visualizations for few-shot segmentation obtained via our method, associated with Table~3 in the main paper. Following the prior work~\cite{partslip_2023}, 8-shot labeled shapes are utilized to carry the few-shot segmentation. From the figure, it is evident that our method successfully achieves satisfactory segmentation results.

\subsection{Convergence curves}
In the ablation studies presented in Table~5, we compare two approaches used in the 3D encoder of our student network. First, we employ a pre-trained PointM2AE~\cite{pointm2aezhang2023} backbone, freeze the weights and only update the learnable parameters in the student network's distillation head. Second, we utilize a PointM2AE backbone with its weights initialized by PyTorch~\cite{pytorch_2019} default initialization and set them to be learnable, together with the parameters in the distillation head. From the table, we observe comparable results between both settings (see rows (5) and (7) for the first and second approaches respectively). 

\begin{figure}[t]
  \centering
  \includegraphics[keepaspectratio, width=0.98\columnwidth]{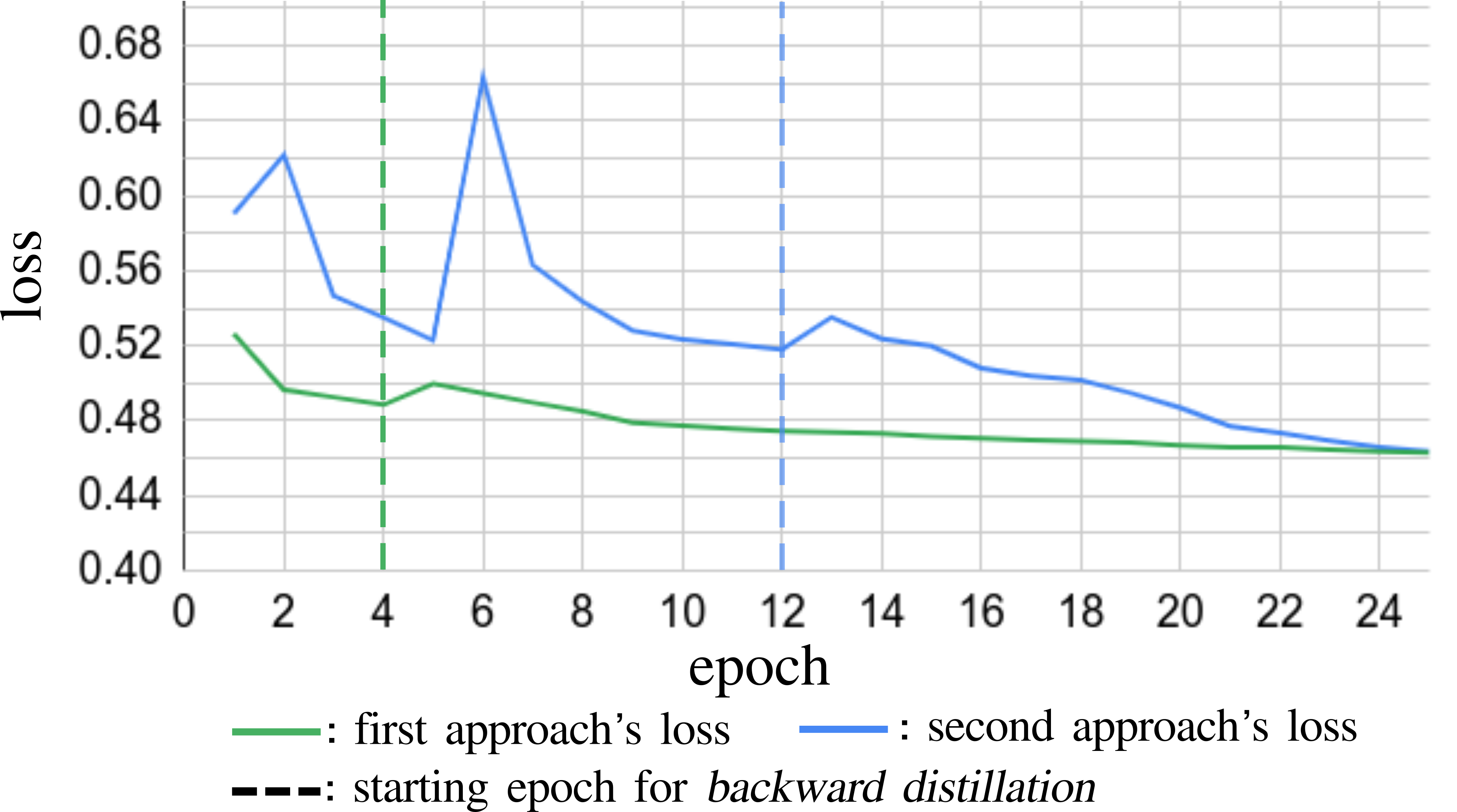}
  \caption{Convergence curves of our method's losses during optimization epochs. While the first approach employs a pre-trained PointM2AE~\cite{pointm2aezhang2023} model and freezes its weights, the second approach initializes the Point2MAE's weights from scratch and sets them to be learnable.\vspace{5px}}
  \label{fig:converge_curve}
\end{figure}

We then visualize the convergence curves in both settings, as depicted in Figure~\ref{fig:converge_curve}. From the figure, it can be seen that the loss in the first approach converges significantly faster than in the second approach.  As a result, the first approach also starts to perform \emph{backward distillation} in a substantially earlier epoch than the second one.

\subsection{2D rendering from various shape types}
\begin{figure}[t]
  \centering
  \includegraphics[keepaspectratio, width=0.98\columnwidth]{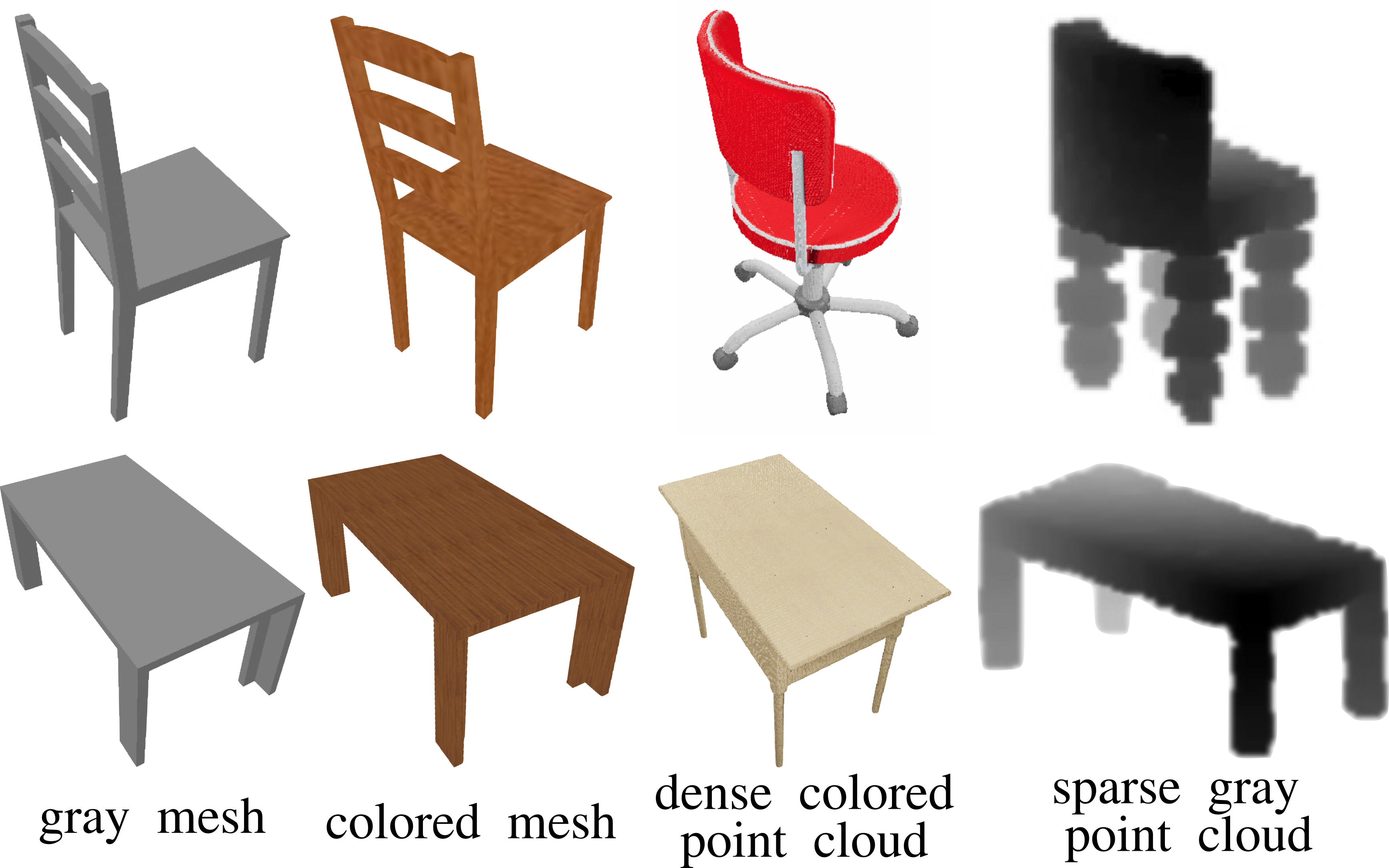}
  \caption{Several examples of 2D images rendered from various shape types.
  }
  \label{fig:rendering2d}
\end{figure}

We present several 2D rendering images from various shape types, including (i) gray mesh, (ii) colored mesh, (iii) dense colored point cloud, and (iv) sparse gray point cloud, which can be seen in Figure~\ref{fig:rendering2d}. While PartSLIP~\cite{partslip_2023} renders the multi-view images using type (iii),  SATR~\cite{satr_2023} uses type (i). As for PointCLIP~\cite{pointclipv1_2022} and PointCLIPv2~\cite{pointclipv2_2023}, they use type (iv) to render their multi-view images.

{
    \small
    \bibliographystyle{ieeenat_fullname}
    \bibliography{main}
}


\end{document}